\documentclass[unnumsec,webpdf,contemporary,large]{oup-authoring-template}
\usepackage{caption}

\usepackage[utf8]{inputenc}
\usepackage{graphicx}%
\usepackage{multirow}%
\usepackage{amsmath,amssymb,amsfonts}%
\usepackage{amsthm}%
\usepackage{mathrsfs}%
\usepackage{xcolor}%
\usepackage{textcomp}%
\usepackage{manyfoot}%
\usepackage{siunitx}
\usepackage{newunicodechar}
\newunicodechar{−}{$-$}
\usepackage{booktabs}%
\usepackage{algorithm}%
\usepackage{algorithmicx}%
\usepackage{algpseudocode}%
\usepackage{listings}%
\usepackage{setspace}%
\usepackage{forloop}%
\usepackage{pgffor}%
\usepackage{epsfig}%
\usepackage{epstopdf}%
\usepackage{wrapfig}%
\usepackage[normalem]{ulem}%
\usepackage{cancel}%
\usepackage{hyperref}%
\usepackage{natbib}
\usepackage{makecell}
\usepackage{array}
\usepackage{rotating}
\usepackage[T1]{fontenc}

\graphicspath{{Fig/}}

\theoremstyle{thmstyleone}%
\theoremstyle{thmstyletwo}%
\theoremstyle{thmstylethree}%

\begin{document}


\firstpage{1}
\title[Short Article Title]{Computational Methods and Challenges in Cell-Free DNA Analysis for Multi-Cancer Early Detection}

\author[1,$\ast$]{Nicko Starkey\ORCID{0009-0009-6390-9354}}
\author[2,$\#$]{Marcin W. Wojewodzic\ORCID{0000-0003-2501-5201}}
\author[1,$\#$]{Krzysztof Rzecki\ORCID{0000-0002-6834-2344}}

\address[1]{\orgdiv{Faculty of Electrical Engineering, Automatics, Computer Science, and Biomedical Engineering}, \orgname{\\AGH University of Krakow}, \orgaddress{\street{Adam Mickiewicz Ave. 30}, \postcode{30-059}, \state{Krakow}, \country{Poland}}}
\address[2]{\orgdiv{Department of Research}, \orgname{Cancer Registry, Norwegian Institute of Public Health}, \orgaddress{\street{Postbox 0379}, \postcode{0379}, \state{Oslo}, \country{Norway}}}

\corresp[$\ast$]{Corresponding authors. \href{email:krz@agh.edu.pl}{krz@agh.edu.pl}}

\received{Date}{0}{Year}
\revised{Date}{0}{Year}
\accepted{Date}{0}{Year}

\abstract{Cell-free DNA (cfDNA) is a promising avenue for non-invasive multicancer early detection (MCED), in that, it can enable multiple cancer detection simultaneously from a single blood draw, with particular sensitivity to cancers that currently lack established screening programs. Here we review the computational methods developed between 2022 and 2025 for cfDNA-based MCED. We focus on how fragmentomics and epigenetic features are extracted and analyzed to detect cancer at early stages. We first briefly outline the biological basis of cfDNA signals, then review classical statistical and machine learning approaches alongside deep learning frameworks including autoencoder-based models. For each method we discuss biological interpretability, validation strategy, and readiness for clinical integration. Furthermore, we categorize the current challenges into technical, computational, and methodological while outlining open problems in the field. This review shows that multimodal ensemble approaches have the strongest promise for clinical integration and the highest readiness. However, for better assessment of future work and side-by-side comparison, standardization of evaluation protocols and reporting results will be crucial.} 




\makeatletter
\renewcommand{\ps@opening}{%
  \def\@oddhead{\thepage\hfil}%
  \def\@evenhead{\thepage\hfil}%
  \let\@oddfoot\@empty
  \let\@evenfoot\@empty
}
\makeatother

\maketitle

\makeatletter
\renewcommand{\ps@headings}{%
  \def\@oddhead{\thepage\hfil}%
  \def\@evenhead{\thepage\hfil}%
  \let\@oddfoot\@empty
  \let\@evenfoot\@empty
}
\pagestyle{headings}
\makeatother

\section{Introduction}\label{sec_intro}
Early detection (ED) of cancer is vital for increasing the survival rate of patients and stratification for targeted therapies. Multiple studies have shown that the survival rate for cancers, such as ovarian, breast, or colorectal, increases from 18--40\% to 90--99\% if cancer is detected in its initial stage compared to its last \cite{imai2025transforming}.

Conventional screening methods, such as Pap smears, biopsies, and imaging (e.g. mammography), have improved monitoring, but remain costly, often invasive, prone to false positives, and insensitive to asymptomatic cancers such as pancreatic and ovarian cancers \cite{milner2024technology}. In lung cancer detection using medical imaging, more than 96\% of positive results can be false alarms \cite{bruhm2025genomic}. Some tests, such as Pap smears for cervical cancer detection, can be unreliable due to their dependence on manual visual inspection, making it challenging to thoroughly examine the large number of cells in each sample. \cite{ACS_PapTest}. At the same time, omics methods significantly decreased in cost, providing the possibility of innovative detection (miRNA, methylation, or genomic test for risk stratification).

Multi-cancer early detection (MCED) refers to the identification of cancer biomarkers via a single blood test to detect multiple cancers at once and in the early stages. MCED offers a transformative alternative to early cancer screening, even for cancers that lack early symptoms. Moreover, widespread adoption could significantly reduce cancer treatment costs \cite{imai2025transforming} and increase survival once such a test is implemented cheaply in practice.

Approximately three decades after Mandel discovered that DNA can be found outside cells and in the plasma of human blood \cite{mandel1948acides}, Leon et al. established the first link between cell-free DNA (cfDNA) and cancer by conducting research on 173 cancer cases and 55 healthy controls \cite{leon1977free}. However, it took another decade for liquid biopsy and cfDNA sequencing to be considered a suitable alternative to invasive anomaly detection methods \cite{thierry2016origins}. Since then, cfDNA has emerged as a minimally invasive biomarker with applications in both screening (early detection biomarkers) and prognosis and treatment monitoring (stratification biomarkers) (\cite{Chung2024,Einstein2020,lau2023single,hosoya2025deciphering,van2024cancer}). Furthermore, unlike tissue biopsy, cfDNA can reflect tumor dynamics throughout the body, making it particularly suitable for MCED \cite{zhang2024circulating}. 

In this review, we highlight computational advances (2020--2025) for cfDNA-based MCED. We first outline the biological underpinnings of cfDNA signals and then review statistical, machine learning, and deep learning approaches for early detection and classification. Finally, we discuss existing challenges, limitations, and future directions toward clinical adoption.

\begin{figure*}[htp]
    \centering
    \includegraphics[width=\linewidth, height=0.85\textheight, keepaspectratio]{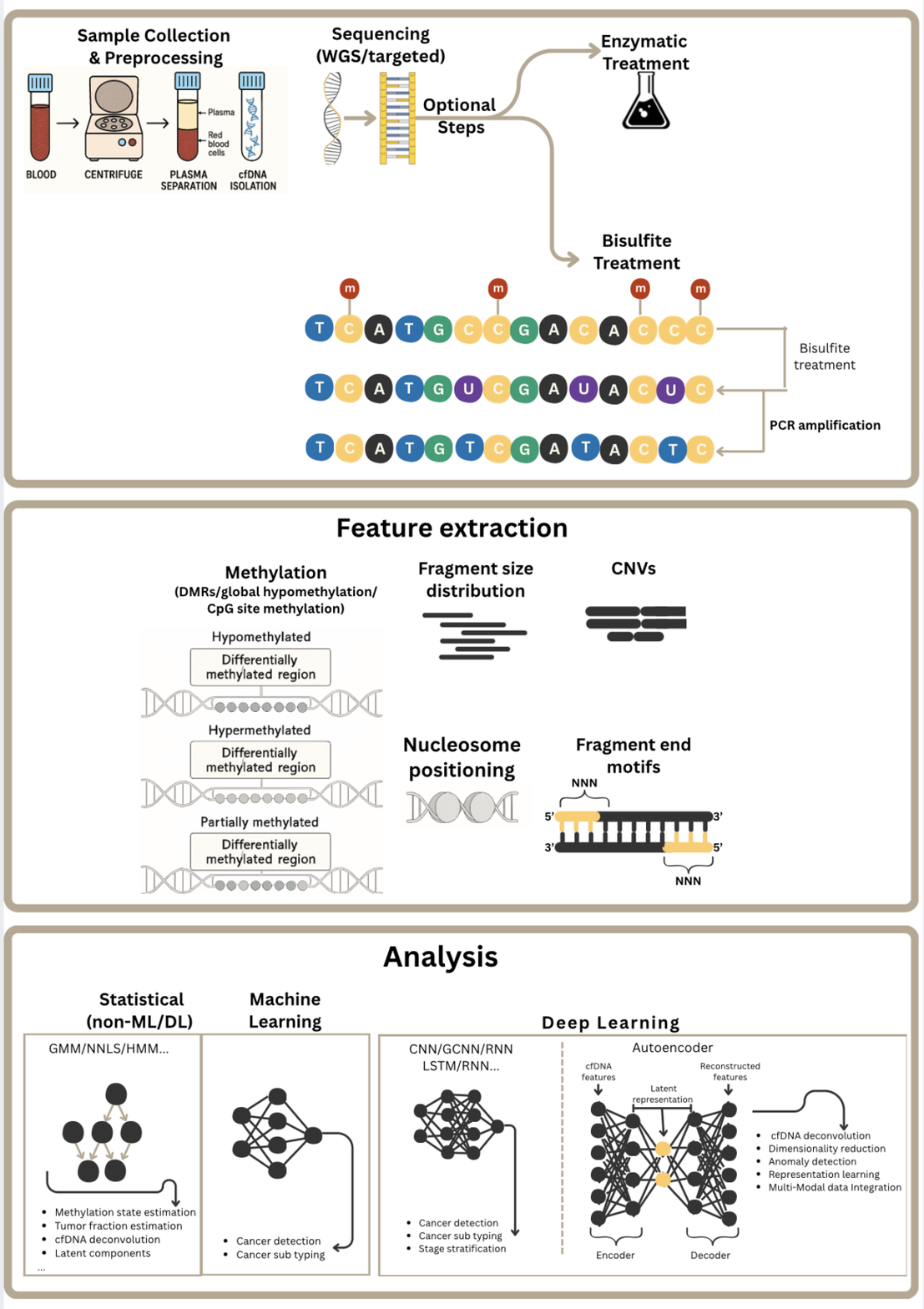}
    \caption{Schematic overview of a cfDNA analysis pipeline. cfDNA from blood is sequenced with bisulfite or enzymatic treatment for methylation profiling. Features are extracted and subsequently analyzed using statistical, machine learning, or deep learning approaches for cancer detection, subtyping, and stage stratification.}
    \label{fig:pipeline}
\end{figure*}

\section{Biological Basis of cfDNA for MCED}
Unlike genomic DNA, cfDNA circulates in blood, saliva, and urine released primarily through apoptosis and necrosis, with apoptosis dominating healthy hematopoietic turnover and necrosis contributing more substantially in cancer and under stressors such as ionizing radiation \cite{giacona1998cell, jahr2001dna, wan2017liquid,heitzer2020cell,siravegna2017integrating}. cfDNA concentration increases with cancer stage and remains elevated even after adjusting for demographic variability (e.g., age, sex, body mass index) \cite{milecki2023preoperative,dihlmann2025exploring,orntoft2021age}. Moreover, cfDNA carries molecular signatures that reflect its tissue of origin (e.g., via methylation marks specific to the tissue of origin). If cfDNA originates from tumor cells, it is termed circulating tumor DNA (ctDNA). Detectable ctDNA has been shown to be found in roughly half of stage I patients and more than four-fifths of stage IV cases \cite{newman2014ultrasensitive, bettegowda2014detection}. 

cfDNA fragmentation is far from random \cite{an2023dna,zhou2022}; specifically, fragments from healthy apoptotic cells peak at $\sim$166 bp (one nucleosome unit), while tumor-derived fragments are typically 10--20 bp shorter \cite{zhu2023circulating,stejskal2023circulating}. The basis for this difference appears to arise from epigenetic and transcriptional changes associated with cancer, in that low DNA methylation leads to a more open chromatin configuration with increased nuclease accessibility that favors cleavage into smaller fragments \cite{an2023dna}. Similarly, elevated transcriptional activity in cancer cells is associated with dynamic chromatin regions that further promote shorter fragment generation \cite{noe2024dna}. In addition to size distribution, nucleosome occupancy patterns and fragment end motifs (FEMs) can preserve information on chromatin structure in tissue of origin, allowing fragmentomics to support not only cancer detection but also classification of tumor type \cite{zhou2022}.

On the other hand, not all fragmentation-based features are universally reliable biomarkers. For example, cfDNA integrity (ratio of long-to-short fragments) has produced conflicting results across studies. Some report a significantly higher concentration and integrity of cfDNA in cancer patients compared to healthy individuals and those with benign tumors \cite{wang2003increased,miao2019clinical,zaher2013cell}, and in cancer patients before surgery or treatment compared to those after \cite{fan2019analysis,wu2019analysis}. In contrast, some measurements show that integrity is higher after tumor removal, and lower in patients with malignant tumors compared to healthy controls \cite{huang2016plasma}. This variability further highlights the need for standardized pre-analytical and analytical protocols before cfDNA integrity can be considered a robust clinical biomarker. 

Another feature of cfDNA that helps distinguish cancer patients from the healthy group is its methylation pattern due to cancer-associated DNA methylation changes that occur early in tumor formation \cite{xu2017circulating}. The predominant pattern is global hypomethylation (lower methylation level than that of the control group), often concentrated within large partially methylated domains (PMDs) \cite{katsman2022detecting}; yet this
coexists with focal hypermethylation at specific loci; for example, in one ovarian cancer study, only 12 of 536 differentially methylated regions (DMRs) were hypomethylated in cancer patients relative to controls \cite{terp2025genome}. Interpretation of cfDNA methylation therefore requires accounting for tumor fraction, genomic region, and assay design. Together, fragmentomic and methylomic features form the biological basis for the computational methods reviewed in Section~\ref{sec:methods} (Fig. \ref{fig:pipeline}).

\section{Computational Methods for cfDNA-Based MCED}\label{sec:methods}

\begin{figure}
    \centering
    \includegraphics[width=1\linewidth]{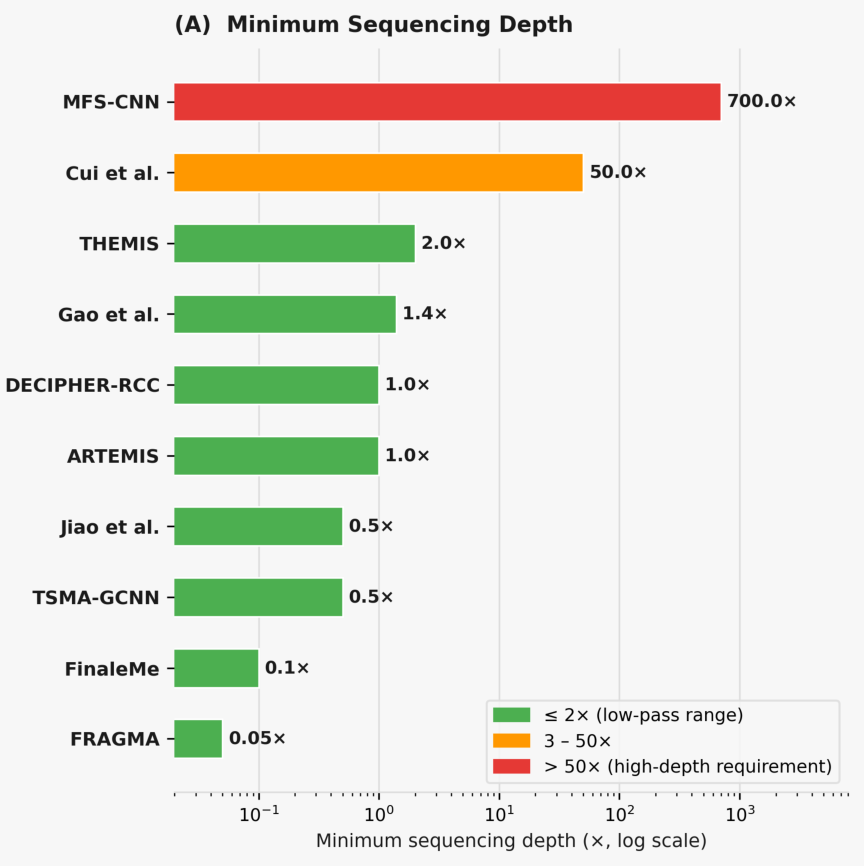}
    \caption{Ranking of methods according to minimum sequencing depth requirement, i.e. the depth at which each method remained performant; sorted ascendingly (log scale).}
    \label{fig:min_seq_depths}
\end{figure}

\begin{figure}
    \centering
    \includegraphics[width=1\linewidth]{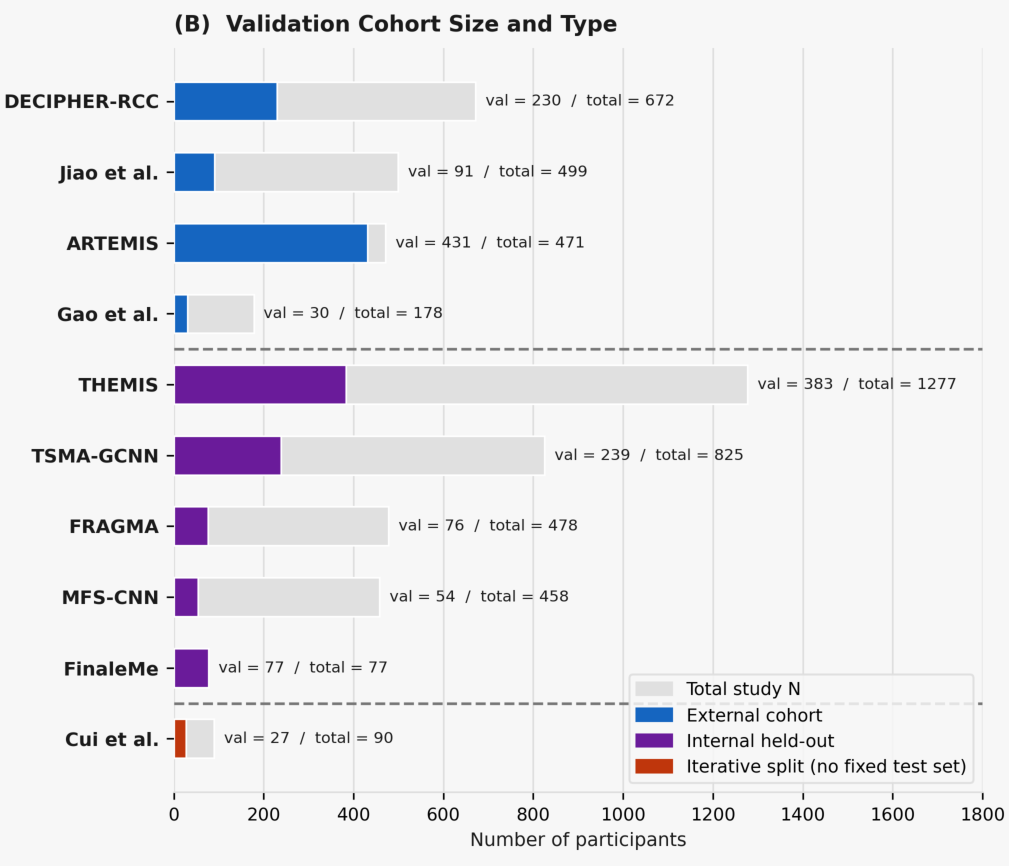}
    \caption{Validation cohort size and type for each method, sorted within category by total study $N$ descending. \textit{External cohort}: independently recruited participants from a separate institution. \textit{Internal held-out}: a fixed, pre-specified split never used for training or tuning; for FinaleMe, this is a purpose-generated matched ULP benchmark set rather than a split of the training set. \textit{Iterative split}: repeated random 70/30 resampling with convergence stopping; no fixed held-out set; $N = 27$ is the mean split size.}
    \label{fig:cohort_sizes}
\end{figure}

Extracting tumor-relevant patterns from the high background of non-tumor DNA in the cfDNA pool \cite{sharma2022computational} requires the development of sensitive computational tools, especially when detecting cancer in its early stages. With the help of next generation sequencing (NGS), machine/deep learning and statistical methods have played a significant role in detecting cancer-associated epigenetic modifications in cfDNA, allowing cancer detection, tissue-of-origin (TOO) identification and disease stage stratification \cite{tsui2025artificial,huang2019bioinformatics}. We reviewed 10 computational methods developed between 2022--2025 for cfDNA-based early cancer detection are reviewed in detail and categorized into two groups based on whether they employ deep learning in their framework. A comprehensive overview of studies and their reported results is provided in Tables \ref{tab:comp_methods}--\ref{tab:per_stage_sens_spec}. The performance metrics shown correspond to the highest level of validation reported in each study prioritizing results from external validation cohorts to better reflect generalizability and the potential for clinical applicability. When external validation was not available, results from held-out internal validation sets are instead included. Furthermore, to maintain consistency with the scope of this review, only values corresponding to early-stage disease are presented. Studies lacking sufficient validation details or those focused on advanced stages were excluded from these comparisons. The ranking of the methods according to their sequencing depth requirement is shown in Fig. \ref{fig:min_seq_depths} and the cohort types and sizes are illustrated in Fig. \ref{fig:cohort_sizes}.

\begin{table*}
\centering
\caption{Computational Methods for cfDNA Analysis Developed Between 2022--2025. ULP: Ultra-Low-Pass; NNLS: Non-negative Least Squares; PLR: Penalized Logistic Regression; GBM: Gradient Boosting Machine; GLM: Generalized Linear Model; WGEM-seq: Whole-genome Enzymatic Methyl-seq}\label{tab:comp_methods}
\begin{tabular}{@{}p{2.2cm}p{3.2cm}p{2.2cm}p{3.5cm}p{2.5cm}p{3cm}@{}}
\toprule
\makecell[l]{\textbf{Method/}\\\textbf{Author}} &
\textbf{Biological Features} &
\textbf{Data Type} &
\textbf{Computational Model} &
\textbf{Task} &
\textbf{Cancer Type(s)} \\
\midrule

FRAGMA \cite{zhou2022}
& Allele-specific differences; tissue-specific cleavage patterns
& WGBS; WGS
& CNN (CpG methylation prediction); SVM (cancer detection)
& Methylation inference; TOO identification; cancer detection
& HCC, NPC \\
\midrule

FinaleMe \cite{liu2024finaleme}
& Fragment length; normalized CpG coverage; CpG distance to fragment center
& WGS; WGBS; ULP-WGS; ULP-WGBS
& Non-homogeneous HMM with GMM initialization
& Methylation status estimation; TOO estimation
& Prostate, breast \\
\midrule

TSMA-GCNN \cite{nguyen2024tissue}
& TSMA deconvolution scores; genome-wide methylation density (GWMD)
& WGBS
& NNLS deconvolution; GCNN
& TOO identification
& Breast, colorectal, gastric, liver, lung \\
\midrule

Gao et al.\ \cite{gao2022whole}
& ctDNA DMRs
& WGBS
& Random forest (feature selection); LASSO logistic regression (detection \& classification)
& Cancer detection; subtype classification
& Breast (early/advanced; ER status) \\
\midrule

MFS-CNN \cite{kim2024deep}
& cfDNA DMRs; fragment size
& WGEM-seq
& CNN
& Cancer detection
& Lung \\
\midrule

ARTEMIS \cite{annapragada2024genome}
& Genome-wide repeat landscapes; cfDNA fragmentation profiles
& WGS
& PLR; GBM; ensemble model
& Cancer detection; TOO identification; disease monitoring
& Lung, liver, breast, colorectal, ovarian, gastric, head/neck, bladder, cervical, thyroid, prostate \\
\midrule

Cui et al.\ \cite{cui2024prediction}
& Cleavage profiles
& WGS; WGBS (ground truth)
& XGBoost (CpG methylation prediction)
& Methylation status inference
& HCC, lung, colorectal \\
\midrule

THEMIS \cite{bie2023multimodal}
& Methylated fragment ratio; fragment size \& end motifs; copy number alterations
& Shallow WMS
& Ensemble ML 
& Cancer detection; TOO identification
& Breast, colorectal, esophageal, liver, lung, pancreatic, gastric \\
\midrule

DECIPHER-RCC \cite{peng2025early}
& CNVs; fragment size distribution; nucleosome footprint
& Low-pass WGS
& Stacked ensemble (GLM, random forest, DL, XGBoost)
& Cancer detection
& Renal cell carcinoma \\
\midrule

Jiao et al.\ \cite{jiao2024leveraging}
& Fragment size coverage \& distribution; nucleosome positioning; CNVs
& WGS
& Stacked ensemble (GLM, random forest, DL, XGBoost)
& Cancer detection
& Esophageal squamous cell carcinoma \\

\botrule
\end{tabular}
\end{table*}


\begin{table*}
\centering
\caption{Performance metrics and key properties of reviewed methods. Seq.\ Depth and Min.\ TF reflect the lowest values reported at which the method remained performant, unless otherwise is indicated. TF: Tumor Fraction; LOD: Limit of Detection; PPV: Positive Predictive Value; CI: Confidence Interval. N/R: not reported.}\label{tab:comp_methods_metrics}
\begin{tabular}{@{}p{2.2cm}p{2.5cm}p{2.9cm}p{2cm}p{2.2cm}p{4.5cm}@{}}
\toprule
\makecell[l]{\textbf{Method/}\\\textbf{Author}} &
\textbf{AUC} &
\textbf{Accuracy} &
\textbf{Seq.\ Depth} &
\textbf{Min.\ TF} &
\textbf{Notes} \\
\midrule

FRAGMA \cite{zhou2022}
& CpG methylation: 0.93; HCC detection: 0.98
& N/R
& $0.05$--$0.5\times$ 
& 2.4\% (early stage)
& PPV boost to 26.8\%; performance stable at shallow depths \\ 
\midrule

FinaleMe \cite{liu2024finaleme}
& CpG methylation: 0.91; TOO: above $0.86$
& N/R
& ${\sim}0.1\times$ (ULP-WGS) 
& N/R
& Reliable in CpG-rich regions; performance drops in CpG-sparse loci \\
\midrule

TSMA-GCNN \cite{nguyen2024tissue}
& N/R
& TSMA deconv: 78\%; TOO (GCNN): 69\%
& $0.5\times$ (TOO); $5$--$15\times$ (TSMA construction)
& 0.01\% (deconvolution range: 0.01--25\%)
& Strong deconvolution at TF~${\geq}10\%$ \\
\midrule

Gao et al.\ \cite{gao2022whole}
& BC detection: 0.967 (early); ER status: 0.909
& N/R
& $1.4-3.7\times$ (samples mean depth range)
& N/R
& DMRs enriched in introns/intergenic regions; consistent plasma-tissue methylation trends \\
\midrule

MFS-CNN \cite{kim2024deep}
& 0.67--0.87 (across region types)
& 81.5\%
& $700\times$
& 0.1\% (at 80\% spec); 1\% (at 98\% spec)
& Better performance in hypomethylated vs.\ hypermethylated regions; SCC outperforms ADC \\
\midrule

ARTEMIS-DELFI \cite{annapragada2024genome}
& All cancers (tissue): 0.96; lung: 0.91; liver: 0.90
& TOO (gDNA): 78\%; TOO (cfDNA): 68\%
& $1\times$ 
& N/R
& Survival correlation ($P{<}0.001$); consistent across sequencing depths and collection batches \\
\midrule

Cui et al.\ \cite{cui2024prediction}
& Healthy: 0.898; cancer: 0.827--0.959
& N/R
& at least $50\times$ to train 
& N/R
& No held-out test set; potential optimistic bias in reported AUCs \\
\midrule

THEMIS \cite{bie2023multimodal}
& Cancer detection: 0.966
& TOO: 54\% (95\% CI: 46--62\%); grouped: 65\% (95\% CI: 57--72\%)
& ${\sim}2\times$ 
& ${<}0.1\%$
& Multimodal; robust against batch effects; minimal overfitting \\
\midrule

DECIPHER-RCC \cite{peng2025early}
& Val.: 0.966; external: 0.952
& N/R
& $1\times$ 
& 0.05\% (at 70\% spec); 0.1\% (at 95\% spec)
& Robust across RCC histological subtypes and stages \\
\midrule

Jiao et al.\ \cite{jiao2024leveraging}
& Val.: ${\sim}0.99$
& Val.: 0.95; external: 0.91
& $0.5\times$ 
& N/R
& Stable performance even at $0.5\times$ depth; most rigorous covariate matching among reviewed methods \\

\botrule
\end{tabular}
\end{table*}


\subsection{Inclusion Criteria and Scope}\label{subsec:inclusion_criteria}
The methods reviewed here were identified through literature searches of PubMed using combinations of the terms \textit{cell-free DNA}, \textit{cfDNA}, \textit{multicancer} , \textit{early detection}, \textit{MCED}, \textit{methylation}, \textit{fragmentomics}, \textit{whole-genome sequencing}, \textit{WGS}, \textit{deconvolution}, \textit{decomposition}, \textit{machine learning}, \textit{deep learning}, and \textit{autoencoder}. The focus was on publications from 2022--2025 that presented the latest computational methods available to the field. 

Computational methods were included if they were based on plasma cfDNA and used whole-genome or whole-methylome sequencing, incorporated machine learning or deep learning for cancer detection, tissue-of-origin (TOO) identification or methylation inference, and if they reported quantitative performance metrics that allowed evaluation. Methods relying solely on targeted mutation panels, tissue-only datasets without cfDNA validation, or lacking sufficient methodological detail for critical assessment were excluded. Applying these criteria, 10 methods were identified for a detailed review.

Several well-known methods predate the 2022 cutoff and are not reviewed in full here. Among these methods, the pioneers are DELFI \cite{cristiano2019genome}, which established genome-wide cfDNA fragmentomic profiling, and CancerSEEK \cite{Cohen2018}, which demonstrated multi-analyte MCED, combining protein biomarkers and mutation detection. These methods form an important methodological precedent to several of the ensemble frameworks reviewed here.

\subsection{Classical Approaches}\label{classical_approaches}
Classical methods, including statistical analysis and traditional machine learning algorithms, have played a crucial role in extracting meaningful signals from the cfDNA pool. These methods employ DNA methylation and fragmentation patterns for cancer detection, TOO identification, and cancer subtyping, often with lower computational demands and greater interpretability than deep learning models \cite{aref2025dna,zhang2025deep}. 

Gao et al. \cite{gao2022whole} developed a WGBS-based method to detect breast cancer and classify its estrogen receptor (ER) status using ctDNA methylation profiles. Their pipeline achieved 87\% sensitivity with 100\% specificity (cancer stage not stratified), identifying 15 DMRs for early and advanced breast cancer detection, and 12 DMRs for the classification of ER status. Although their method showed superior performance over previous results with reduced representation bisulfite sequencing RRBS data for gene distribution, the approach relies heavily on high-quality ctDNA extraction, which may constrain its clinical utility in low-input or early-stage settings. The study was restricted to female participants for breast cancer and was stratified for ER subtype cohorts (30 ER$-$ and 30 ER$+$ patients). However, cancer patients were approximately 6--11 years older than controls on average, and postmenopausal status was over-represented among cancer patients relative to controls. Given that cfDNA methylation is known to vary with age \cite{chen2025whole}, and postmenopausal status is strongly associated with ER$+$ disease (approximately 80\% of ER$+$ breast tumors that occur in postmenopausal women \cite{majumder2017post}), there is a concern that the model may be partly learning signals related to age or menopause rather than cancer specific patterns. 

In 2024, Liu et al. \cite{liu2024finaleme} introduced FinaleMe, a non-homogeneous hidden Markov model (HMM), which predicts CpGs methylation in fragments with $\geq$5 CpGs with an area under the curve (AUC) of $0.91$. They used the length of the cfDNA fragment, the midpoint distance, and the normalized coverage of CpG in the reference genome as inputs to a Gaussian mixture model (GMM) to initialize the starting methylation state of each CpG site in each cfDNA fragment by modeling the distribution, using deep and shallow WGS and WGBS cfDNA samples. FinaleMe was evaluated primarily in breast and prostate cohorts, but also estimates TOO, which serves as the potential to be applicable to multicancer settings. Covariate control is not directly applicable here due to the fact that the reported performance metrics reflect the prediction of CpG methylation benchmarked against a WGBS ground truth on the same samples, not a case-control classification where demographic imbalance could inflate the results. Similarly, the TOO estimate is not evaluated against a held-out case-control cohort in a manner that would make age or sex differences a meaningful confounder.

Using fragmentation patterns in cfDNA, Cui et al. \cite{cui2024prediction} used an extreme gradient boosting algorithm (XGBoost) to predict the methylation status of single CpG sites. The authors utilized WGBS methylation data as ground truth, and WGS to spot cleavage profiles that can be fed to the model as cancer biomarkers for MCED methylation predictions. The study was conducted on 90 samples in total, from healthy individuals and patients with hepatocellular carcinoma (HCC), lung, and colorectal cancers. Although this method achieved high performance metrics (more than 82\%, Tables \ref{tab:comp_methods_metrics} and \ref{tab:aggregate_sens_spec}) across healthy and cancerous samples, the results were not stratified by cancer stage, no clinical covariate control was conducted, nor was demographic information of participants reported. Thus, these results should be interpreted with caution. Additionally, this method requires a substantially high sequencing depth (Fig. \ref{fig:min_seq_depths}), which position it further away from clinical application due to costs.

THEMIS (THorough Epigenetic Marker Integration Solution \cite{bie2023multimodal}), on the other hand, reduces the cost of sequencing in their analysis by using shallow whole methylome sequencing (WMS). WMS, as demonstrated in the study through Pearson correlation analysis, reliably mirrors WGS when extracting fragmentation profiles from cfDNA data. THEMIS integrated methylated fragment ratio (MFR), fragment size index (FSI), chromosomal aneuploidy (CAFF), and fragment end motif as features, designed to jointly capture epigenetic and fragmentomic alterations associated with ctDNA. Then a regularized linear regression model is utilized to combine all the resulting predictions together with the aneuploidy levels to achieve a unified cancer detection probability. It was tested on a large multicancer cohort of 1277 samples in total and achieved a near perfect AUC in cancer detection ($0.97$). However, the method was validated using an internal set, and batch effects were ruled out using visual clustering methods. The authors acknowledge that there was no specific covariate matching between the cancer and control groups, including age information with the cancer patients on average 7--11 years older than controls. Given the performance reported and the cohort size, further development of this method may demonstrate its candidacy for clinical applicability.

ARTEMIS (analysis of repeat elements in disease \cite{annapragada2024genome}) is another framework that features an ensemble learning framework. It distinguishes ctDNA from normal plasma by investigating the distributions of repetitive elements throughout the genome using $100$bp--long reads. Although the framework was originally applied to both tissue and plasma samples, its implementation of cfDNA focuses on summarizing large-scale repeat-sequence variation linked to genomic instability and epigenetic regulation. In this setting, ARTEMIS derives quantitative descriptors of repeat element composition, augments them with chromatin-context features from repeat-dense loci, and incorporates fragmentomics to enhance performance, achieving AUCs of up to 0.91 in prospective lung and liver cancer screening cohorts \cite{annapragada2024genome}. The researchers thoroughly assessed and controlled for variables such as sex, race, tumor type, and cancer-associated gene size to ensure that these do not confound the cancer signal being detected.

\subsection{Deep Learning Approaches}\label{subsec:dl_approaches}
Deep learning (DL) models have demonstrated great ability in learning high-dimensional, nonlinear representation of complex biomedical data in recent years \cite{yi2022graph,zhong2016overview}. This gives them an edge over traditional machine learning (ML) models and is driving progress in developing advanced tools for methylation analysis, particularly in MCED \cite{sahoo2025artificial}. 

DL models can extract latent features, including methylation and fragmentomic patterns, that can be used in cancer detection, classification, and TOO identification \cite{tsui2025artificial,li2021dismir,li2023comprehensive}. However, in the case where the number of samples is small ($<10\times$ the number of parameters), these models are generally prone to overfitting \cite{tsui2025artificial} and require explainability tools to ensure that their output is aligned with domain knowledge \cite{zhang2025deep}. In the following, some of the recent DL models developed for cfDNA-based MCED are further detailed. 

Nguyen et al. \cite{nguyen2024tissue} developed a framework (denoted TSMA-GCNN throughout this review) to predict TOO in low-depth cfDNA samples using WGBS data from $5$ tumor types coupled with white blood cells (WBCs); initially they created a tumor-specific methylation atlas (TSMA). However, given the high proportion of DNA fragments from WBC in the cfDNA pool, TSMA alone proved to be insufficient for TOO detection. The proposed solution was to combine the deconvolution scores from TSMA with genome-wide methylation density (GWMD) features and use this set of features as input to a GCNN. No individual-level demographic matching was reported between cancer and control cases. Although TSMA-GCNN remained feasible for tumor fraction (TF) of $0.01$\% and the model was externally validated, the performance is only acceptable when TF $\geq 10\%$ (Table \ref{tab:comp_methods_metrics}), and the article does not report sensitivity and specificity, nor does it mention any clinical demographic matching or covariate correction.  

DL has also been increasingly applied to infer the status and patterns of DNA methylation from fragmentomic features. An example is FRAGMA \cite{zhou2022}, a CNN-based framework that interprets cleavage frequency patterns within short genomic windows surrounding CpG sites to distinguish hypo- and hypermethylated regions. In addition to achieving a high single-CpG AUC of $0.93$, FRAGMA offers this methylation analysis without relying on chemical treatments (e.g., bisulfite conversion). This is particularly significant, as it has been reported that DNA undergoes a degradation of 84--96\% during such treatments \cite{grunau2001bisulfite}. As a result, employing DL-based methods to eliminate bisulfite treatment, and therefore allowing the use of more intact DNA, is a promising direction in methylation analysis \cite{liu2019detection, hu2023prediction, ahsan2024signal}. As this method performs a case-control classification, where age and sex differences between HCC patients and controls could plausibly confound the signal, it is necessary to report or adjust for these demographic variables. However, nuclease activity (DNASE1L3) was the only biological confounder explicitly acknowledged. Additionally, similar to Gao et al. and Cui et al., this study does not report stage-stratified performance results \ref{tab:aggregate_sens_spec}.

The use of DL and ML in tandem has also been tested in cfDNA-based early cancer detection studies. Two representative examples use an identical stacked ensemble strategy, combining gradient boost and generalized linear models, random forests, and a DL component, which are trained on comparable fragmentomic features of the cfDNA. The first was developed by Jiao et al. \cite{jiao2024leveraging} for esophageal squamous cell carcinoma (ESCC), used fragmentation size coverage and size distribution, nucleosome footprint, and the CNV, extracted from the WGS data. They trained the models on each feature type separately before combining the top performers into an ensemble. The framework was validated on two separate validation datasets (141 ESCC and 151 controls, in total) and demonstrated high sensitivity, specificity, and robustness, maintaining stable performance even when sequencing depth was lowered from $4\times$ to $0.5\times$. Among the methods reviewed, Jiao et al. adopted the most rigorous approach to cancer-control clinical covariate matching. They specifically addressed age and sex across all cohorts---the balance statistically confirmed (p-value $>0.05$)---as well as smoking and drinking status. However, there is no sensitivity and specificity reported. The second framework was developed by Peng et al. (DECIPHER-RCC) \cite{peng2025early}, which applied the same ensemble approach to renal cell carcinoma (RCC), training on WGS cfDNA data to extract CNV, fragment size distribution and nucleosome footprint features. The method retained strong detection capability at sequencing depths as low as 1–3$\times$ and tumor fractions down to 0.1\% at 95\% detection (0.05\% at 70\%). These studies illustrate the adaptability of a shared DL-ML design across cancer types. Peng et al. reported comparable demographics in their training, validation, and external cohorts. The median ages ranged from 48 to 56 years, and the sex distributions were generally similar between groups (approximately 50 to 70\% male throughout).

Some approaches have fused fragmentomics and methylomics data to enhance analysis performance.  Using 366 lung cancer-specific biomarkers, Kim et al. \cite{kim2024deep} designed a targeted enzymatic methyl-seq panel of 142 lung cancer samples and 56 controls. The combination of methylation and fragment size features derived from this panel was passed to a CNN as a 2D vector. The model, with an AUC of$ 0.87$ was sensitive enough to detect tumor fractions as low as 1\%  and 0.1\% at specificities of 98\% and 80\%, respectively. According to the age and sex information reported for both cancer and non-cancer groups, a notable demographic imbalance seem to have remained unaccounted for in the analysis, in that cancer patients had a mean age of 66 years and were predominantly male (76\%), while healthy controls had a mean age of 32 years and were predominantly female (59\%).

\subsubsection{Autoencoders and cfDNA}
Autoencoders (AEs) are neural network models that learn compact latent representations by reconstructing their input data. This makes them especially useful for cfDNA analysis, where high-dimensional methylation or fragmentomic features are sparse and noisy. An AE compresses the cfDNA profiles through a narrow bottleneck, termed \textit{the latent space}; this is done by an encoder. This latent space is passed to a decoder that reconstructs the input. This reconstruction is only considered successful if the latent features capture the essence of the original features, i.e. the real signal \cite{zhai2018autoencoder}. In this context, the encoder is intended to provide a biologically meaningful feature space, while the decoder regularizes against noise and enforces consistency with the underlying molecular patterns \cite{yang2025efficient, jackson2023deconvolution,macias2020autoencoded}. As a result, AE-based pipelines are increasingly applied to cfDNA along two lines of work, namely, learning a representation that is then fed to a classifier, and resolving the cell composition of the cfDNA pool through signal deconvolution. AE-based studies for the context of cfDNA, albeit limited at the time of writing this, are promising ways to capture complex patterns associated with cancer.

The first line employs the AE as a feature generator. Many methods summarize methylation levels within genomic regions via a scalar. This discards methylation signatures in specific locations, limiting the detection of early-stage markers. DeepMeth \cite{cai2022noninvasive} addresses this by learning a continuous representation of each region and then passing the compressed feature space to a multi-layer perceptron, a CNN, and a recurrent neural network for early-stage lung cancer detection. Its encoder comprises four residual blocks and treats each region as a one-channel image and returns a vectorized region, and the decoder reconstructs the region from the latent space using a reversed residual network. They validated DeepMeth on $273$ malignant and $151$ benign samples from $14$ hospitals, achieving AUC of $\sim0.79$--$0.81$. DeepMeth is a clear demonstration that auto-encoded features at the region level transfer to the liquid biopsy for early detection. That said, the idea of compressing methylation into encoded features predates its use in cfDNA; for example, Mac\'ias-Garc\'ia et al. \cite{macias2020autoencoded} used an AE with a single hidden layer before survival analysis to compress $450$K methylation arrays from breast tumors.  

The second line addresses cfDNA signal deconvolution, the difficulty of which is due to hidden dependencies and complex patterns that require advanced analytical approaches, and for which deep learning is well-suited. MetDecode \cite{passemiers2024metdecode} treats the cfDNA methylation profile as a weighted mixture of signals from multiple tissues and seeks to recover those weights. It does so through reference-guided matrix factorization, adding coverage-weighted optimization for noisier measurements. What was missing from the reference atlas was learned directly from the input, which, to some extent, compensates for the need for a fixed atlas as a supplement. MetDecode detection is effective for TF as low as 2.88\% and correctly identifies the tissue of origin in 16 of 19 patient samples above this threshold. 

Using simulated data, Jackson et al. \cite{jackson2023deconvolution} developed a deep AE constructed from an encoder that takes a cfDNA methylation matrix as input and returns a vector of tissue contributions, with its decoder using a mean Spearman's correlation loss to reconstruct the original methylation profiles, achieving a Spearman correlation of $0.83$ for variable contributions and $0.965$ for random splits. Their model outperformed a previously developed method, based on non-negative least squares \cite{Moss448142}, by approximately 20\% for variable contributions and 2.5\% for random splits. They then validated their model using a cohort of $21$ HCC samples and $30$ controls, obtaining an AUC of $0.93$ in cancer classification using a random forest. While this method was validated on real data, as the authors hinted, the physiological diversity of the non-cancer cohort raises concerns about the model's generalizability. In case this variability is too high, it might introduce variability that makes it more difficult to determine whether the model truly distinguishes cancer-related patterns or just picks up on unrelated physiological differences.

cfDecon \cite{wang2025cfdecon} deconvolutes at a single read resolution with a multichannel autoencoder that takes cfDNA methylation profiles as a tensor, and captures regional correlations and nonlinear properties of complex cellular mixtures in attempt to accurately estimate cell-type proportions while maintaining interpretability as the activation layers and the biases are eliminated from the decoder, which is explained in \cite{chen2022deep}. cfDecon showed a 33\% enhancement in correlation over non-AE methods CelFiE \cite{caggiano2021comprehensive} and cfSort \cite{li2023comprehensive} on simulated data from normal human methylome \cite{loyfer2023dna}. For disease diagnosis, this method reported 49\% and 40\% accuracy improvements on Amyotrophic lateral sclerosis (ALS) and HCC, respectively, compared to prior methods.

The problem with using a fixed methylation atlas is that it cannot accommodate the biological variability of tumors across individuals and disease stages. Oncoder \cite{yang2025efficient} embeds the reference learning inside the model, where features a decoder that jointly learns tumor and healthy methylation profiles from the training data while the encoder estimates tumor fraction with a beta-distribution penalty keeping the learned profiles biologically grounded. Compared to NNLS, Oncoder demonstrated a relative increase of $\sim16$\% in sensitivity and $\sim37$\% in specificity for HCC detection at its ''earliest stage''.

To date, the number of methods that apply AE specifically to cfDNA for early stage cancer detection is still limited, and broader claims regarding generalizability and clinical utility will require validation in larger and more diverse cohorts.

\begin{table}[!ht]
    \centering
    \caption{Sensitivity and specificity of methods whose original studies reported only non-stage-stratified values or did not provide their definition of the word "early". Methods that reported per-stage breakdowns appear in Table~\ref{tab:per_stage_sens_spec} instead. Values are rounded to the nearest integer.}
    \label{tab:aggregate_sens_spec}
    \small
    \begin{tabular}{lllll}
    \toprule
        \textbf{Method} & \textbf{Sensitivity (\%)} & \textbf{Specificity (\%)} & \textbf{Cancer Type} & \textbf{Stage} \\
    \midrule
        FRAGMA & 80 & 96 & HCC & Early \\
        Gao et al. & 87 & 100 & Breast & I, II, III \\
    \midrule
        \multirow{3}{*}{Cui et al.} & 86 & 100 & HCC & Early \\
                                    & 88 & 82 & Lung & Early \\
                                    & 100  & 78 & Colorectal & ``Early'' \\
    \bottomrule
    \end{tabular}
\end{table}

\begin{table}[!ht]
    \centering
    \caption{Sensitivity and specificity of methods whose original studies reported per-stage values. Methods that reported only aggregate sensitivity appear in Table~\ref{tab:aggregate_sens_spec} instead. Where two sensitivity values are reported, they correspond to Stage I and Stage II, respectively. Values are rounded to the nearest integer.}
    \label{tab:per_stage_sens_spec}
    \small
    \begin{tabular}{llll}
    \toprule
        \textbf{Method} & \textbf{Sensitivity (\%, I ; II)} & \textbf{Specificity (\%)} & \textbf{Cancer Type} \\
    \midrule
        MFS-CNN & 43, 57 & 98 & Lung \\
    \midrule
        \multirow{7}{*}{THEMIS} & 100; 75  & 99 & Breast \\
                                & 67; 71   & 99 & Colorectal \\
                                & 60; 100  & 99 & Esophageal \\
                                & 58; 83   & 99 & Liver \\
                                & 50; 67   & 99 & Lung \\
                                & 88; 73   & 99 & Pancreatic \\
                                & 100; 100 & 99 & Gastric \\
    \midrule
        \multirow{2}{*}{DECIPHER-RCC} & 71; 89 & 95 & RCC \\
                                      & 63; 78 & 98 & RCC \\
    \midrule
        \multirow{2}{*}{ARTEMIS-DELFI} & 47; 38 & 80 & Lung \\
                                       & 62; 44 & 70 & Lung \\
    \bottomrule
    \end{tabular}
\end{table}

\subsection{Overview of Computational Models}
\subsubsection*{Biological Interpretability and Clinical Application} 
Table \ref{tab:interpretability_validation} shows an overview of all the computational models reviewed. The second, third, and fourth columns present biological interpretability, clinical readiness, and the level of demographic matching and covariate control (discussed in detail in Section \ref{sec:methods}), respectively. There are various definitions for the interpretability of a model, none of which is universally agreed upon. However, for the purposes of this study, a model or framework is considered interpretable if i) features, outputs, or mechanisms explicitly linked to biological processes or molecular characteristics (e.g., methylation patterns) provide insight into the underlying biology of cfDNA or cancer, and ii) there is external biological domain knowledge integrated into the design (e.g., through cancer pathways) \cite{wysocka2023systematic}. Similarly, complete validation for clinical application is inherently limited as no predictive or diagnostic model can be confirmed for all future clinical conditions. Therefore, the heterogeneity in performance of these models is an inevitable outcome of contextual variations. Validation of such models requires maintenance, which means a well-documented ongoing validation process in multiple environments \cite{van2023there}. That said, researchers have proposed frameworks and criteria to assess the extent to which a model meets these requirements \cite{nguyen2021appraising}. In this review, a framework or model is considered validated for clinical application if the study provides evidence of testing in a clinical or real-world setting. This includes validation on independent clinical cohorts (e.g. multicenter patient samples, external datasets), performance metrics tied to clinical outcomes, such as sensitivity/specificity for cancer detection, and application to clinically relevant scenarios including early cancer detection, subtype classification, or TOO identification in patient samples. Models tested only in silico or in nonclinical datasets are not considered clinically validated. ARTEMIS and Jiao et al. stand out as the most comprehensively validated frameworks, combining biological interpretability with rigorous covariate control and external validation. On the contrary, methods such as FRAGMA and Cui et al. remain internally validated or not clinically ready, lacking demographic adjustment and external cohort testing. Overall, ensemble-based approaches have the highest promise for clinical applicability.

\begin{table*}[h!]
\centering
\caption{Overview of computational models. BI: Biologically Interpretable; CR: Clinical Readiness; CC: Covariate Control (None = no demographic variables reported or adjusted; Partial = some variables reported but residual imbalances remain unadjusted; Rigorous = key covariates explicitly matched or adjusted with statistical confirmation; N/A = not applicable to the method's primary evaluation design).}
\label{tab:interpretability_validation}
\begin{tabular}{p{1.5cm} p{1cm} p{1.5cm} p{1.5cm} p{3.5cm} p{3.5cm}}
\toprule
\textbf{Model} & \textbf{BI?} & \textbf{CR} & \textbf{CC} & \textbf{Key Strengths} & \textbf{Key Limitations} \\
\midrule
FRAGMA       
& Yes      & Internally Validated  
& None     & Bisulfite-free; feasible seq.\ depth              
& No external validation; DNASE1L3 confounding \\[10pt]

FinaleMe    
& Yes      & Internally Validated            
& N/A      & Bisulfite-free; multiple depths tested             
& CpG-sparse performance drops; no sens/spec \\[10pt]

TSMA-GCNN    
& Partly   & Externally Validated              
& None     & Ultra-low depth; multi-cancer                      
& High TF requirement for TOO; no sens/spec; low accuracy \\[10pt]

Gao et al.   
& Yes      & Externally Validated             
& Partial  & High AUC; minimal input; ddPCR confirmed           
& Age and menopausal status unadjusted; external validation used tissue not cfDNA data; small cohort size for validation \\[10pt]

MFS-CNN      
& Partly   & Internally Validated              
& None     & Detects TF to 0.1\%; panel-based                  
& Large age/sex imbalance unadjusted; low early-stage sensitivity \\[10pt]

ARTEMIS      
& Yes      & Externally Validated              
& Rigorous & Multi-cancer; low-coverage WGS; prospective        
& Low AUC for early lung; germline-variable repeats \\[10pt]

Cui et al.   
& Yes      & Not ready    
& None     & Bisulfite-free; genetic-epigenetic profiling       
& No held-out test set; no external cohort \\[10pt]

THEMIS       
& Yes      & Internally Validated         
& Partial  & Multi-hospital; batch-corrected; low-pass WMS      
& Age gap unadjusted; low TOO accuracy; no external validation \\[10pt]

DECIPHER-RCC 
& Yes      & Clinically Promising         
& Partial  & External validation; robust at shallow depth; comparable demographics           
& Single cancer type; covariate balance not formally tested \\[10pt]

Jiao et al.  
& Yes      & Clinically Promising              
& Rigorous & Robust at shallow depth; statistically confirmed covariate balance 
& Single cancer type (ESCC) \\
\botrule
\end{tabular}
\end{table*}

\subsubsection*{Data Leakage and Overfitting} 
A critical aspect in evaluating computational frameworks lies in how effectively each study mitigates the risks of data leakage, overfitting, or inadvertent information transfer between training and test sets. 

Nguyen et al. (TSMA-GCNN) employed a semi-supervised approach in which the model uses the structure of both labeled and unlabeled data during training to make predictions specifically for the given test instances. For TOO identification, they reserved an independent test set from their discovery cohort. Since during learning the labels of the test nodes remained unseen, no test information influenced the model during training. In FinaleMe, Liu et al. assessed overfitting by training their HMM model on gDNA with no methylation‑-fragmentation relationship. The absence of spurious methylation differences indicated that the model was not memorizing artifacts, reducing the risk of data leakage from training data. Both Gao et al. and Zhou et al. (FRAGMA) used a two-way split with a held-out test sets to evaluate their models. However, Gao et al. went beyond securing unseen data for the testing phase; they also pruned selected features one by one in descending order of their importance scores to help further prevent potential overfitting. Finally, Jiao et al. recruited training, validation, and external cohorts at nonoverlapping time intervals, structurally preventing any leakage from the start. The cutoff was determined exclusively on the training cohort and then applied to the validation and external cohorts without modification. Additionally, a 5-fold cross-validation was employed on the training set for both the individual models and the stacked ensemble, and an ensemble of the top 10 models identified via random grid search was used as the final predictor rather than a single tuned model.

Another way that can potentially limit leakage is to split data iteratively and randomly into training and validation subsets and assess the variation in performance metrics across runs, which was the approach taken by Cui et al. to ensure the model's robustness. However, the study does not specify the use of a held-out test set nor a nested tuning procedure where the parameters are tuned only based on the training subsets. Therefore, this process may inadvertently incorporate tuning feedback from evaluation subsets \cite{cawley2010over}. In contrast, in methods like THEMIS and MFS-CNN, while taking a similar splitting approach, developers alleviated overfitting and data leakage by fully separating model tuning from evaluation. The early stopping in MFS-CNN was triggered by increases in validation loss, while with THEMIS, they stopped the training by monitoring the read count and detecting when the model starts to overfit the extracted features, ensuring the model is only tested on an independent cohort. Furthermore, THEMIS combines bootstrap‑-based splitting and leave‑one-out cross‑-validation with a separate test set, while MFS-CNN uses a straightforward three‑-way split. DECIPHER-RCC and ARTEMIS, too, apply cross-validation but in different ways; DECIPHER-RCC uses a standard 5-fold; however, the model optimization phase in ARTEMIS is isolated from evaluation at every stage by implementing a nested cross-validation, with an inner loop for model selection and an outer loop for an unbiased generalization performance. In addition, ARTEMIS is tested on multiple external cohorts, making it a relatively reliable framework.

\section{Challenges in cfDNA-Based Cancer Detection}
In this section, we first address challenges for which the reviewed methods have already demonstrated mitigation and then discuss open problems for which no satisfactory solution yet exists. In Table \ref{tab:challenges}, these challenges are classified as technical (arising from biological or physical properties of cfDNA that cannot be resolved by changing the study design), computational (arising during data processing and/or modeling) or methodological (arising from experimental protocols, cohort design, or analytical choices).

\begin{table*}[h]
\centering
\caption{Challenges in cfDNA-based MCED}
\label{tab:challenges}
\begin{tabular}{p{2.5cm} p{3cm} p{3.5cm} p{3.5cm}}
\toprule
\textbf{Challenge} & \textbf{Explanation} & \textbf{Significance} & \textbf{\makecell{Mentioned/Proposed\\Mitigation}} \\

\midrule
\multicolumn{4}{c}{\textbf{Technical}} \\
\midrule
\makecell[l]{Scarcity of\\ctDNA \cite{cui2024prediction, gao2022whole}} & Low ctDNA abundance in early-stages challenges sensitive detection & Lowers sensitivity in screening $\Rightarrow$ risks false negatives and ineffective early interventions & Methods preserving DNA integrity; targeted deep-sequencing panels;pre-analytical protocol standardization \\


\midrule
\multicolumn{4}{c}{\textbf{Computational}} \\
\midrule
\makecell[l]{High-Seq Depth\\Requirement \cite{cui2024prediction}} & Needs $>50\times$ WGS depth for CpG cleavage analysis $\Rightarrow$ high sequencing costs & Increases expense, reduces generalizability and equity in cfDNA ED programs & Optimize models for lower-depth WGS; await dropping sequencing costs \\

Low Signal-to-Noise ratio \cite{gao2022whole} & Non-tumor cfDNA background obscures ctDNA signals & Increases false negatives/ positives & Comprehensive filtering using deep-learning algorithms \\

Sens/Spec Limits \cite{gao2022whole} & ED needs high sens/spec with low ctDNA & Risks missed detections or overdiagnosis $\Rightarrow$ limits clinical utility & Optimize deep-learning algorithms; validate with multicenter cohorts \\

\midrule
\multicolumn{4}{c}{\textbf{Methodological}} \\
\midrule

\makecell[l]{Bisulfite Treatment\\Challenges \cite{cui2024prediction}} & Bisulfite treatment is costly, can degrade cfDNA and may alter cleavage signatures & May reduce sensitivity for early-stage cancer screening, introduce bias in detection accuracy, limit accessibility and scalability of cfDNA-based cancer screening in clinical settings & Enzymatic conversion; direct methyl-sequencing on third-generation platforms; utilize cleavage profile-methylation status correlation using WGS assays to avoid bisulfite; further studies on bisulfite treatment's effects\\
&&&\\

Limited Genome-Wide Methods \cite{gao2022whole} & Lack of unbiased genome-wide methylation detection methods & Limits comprehensive biomarker discovery & Improve existing WGBS-based methods for cfDNA-based ED \\

&&&\\

Lack of Standardized Reporting & No consensus on cohort composition, metrics, or validation standards & Prevents cross-study comparison; slows regulatory and clinical translation progress & Adopt minimal reporting standards for liquid biopsy studies; mandate stage-stratified performance metrics \\

&&&\\

\botrule
\end{tabular}
\end{table*}

\subsection{Challenges with Already Proposed Mitigation}
\subsubsection*{High Sequencing Depth Requirements}
Deeper sequencing or highly specific targeted panels is required to reliably detect ctDNA when tumor fractions below 1\%, as signal-to-noise ratio drops due to the large non-tumor cfDNA background \cite{bronkhorst2023changing}. Methods that require depths of $\geq 50\times$ or $700\times$ \cite{cui2024prediction, kim2024deep} present a significant barrier to scalable screening. That said, the methods reviewed in this work collectively demonstrate that high depth is not a universal requirement and multimodal ensemble designs can compensate for depth reduction \cite{peng2025early, jiao2024leveraging, bie2023multimodal, annapragada2024genome, liu2024finaleme}. Figure~\ref{fig:min_seq_depths} illustrates this disparity; minimum sequencing depths span five orders of magnitude across the reviewed methods, from 0.05$\times$ (FRAGMA) to 700$\times$ (MFS-CNN), with the majority operating below 2$\times$.

\subsubsection*{Bisulfite Treatment}
Standard WGBS requires bisulfite conversion, leading to significant DNA degradation, which can impose alteration on fragmentation patterns \cite{grunau2001bisulfite, cui2024prediction, bie2023multimodal}, potentially confounding methods that model fragmentation profiles. The reviewed methods show that bisulfite conversion is not a strict constraint, with enzymatic conversion, fragmentation-based methylation inference, and DL-based bisulfite-free approaches all represented as viable alternatives \cite{kim2024deep, zhou2022, liu2024finaleme, liu2019detection, hu2023prediction, ahsan2024signal}. However, whether these alternatives remain reliable at low tumor fractions and CpG-sparse regions most relevant to early-stage MCED is the outstanding question.

\subsubsection*{Data Sparsity, Missing Coverage, cfDNA Deconvolution}
One of the primary challenges in cfDNA analysis is sparsity, that is, a high rate of missing values across genomic regions. This missingness can arise from technical limitations in sequencing and low DNA yield. Naive imputation of missing data without meticulous consideration of the causes is not a recommended practice in the biological context, as it may bias statistical tests and obscure or distort underlying biological relationships \cite{gorlova2025exposure,krutkin2025impute}. Another fundamental computational barrier is the uneven distribution of biologically informative regions throughout the genome, which can undermine the generalizability and classification accuracy of a model. For example, although FinaleMe \cite{liu2024finaleme} performs strongly in regions rich in CpG, its prediction performance drops significantly in regions with low CpG density, which implicates the reliability of diagnosis and TOO estimation. More broadly, high dimensionality of the cfDNA feature spaces (genome-wide methylation or fragmentation profiles) relative to the small number of clinically annotated samples remains a challenging aspect of training robust models without overfitting. Restricting analysis to high-confidence recurrent regions \cite{gao2022whole}, and strategies such as transfer learning \cite{zhang2025cfmethylpre} and leveraging reference methylation atlases \cite{yang2025efficient} have been shown to mitigate the aforementioned challenges, even though neither fully resolves the challenge in CpG-sparse loci or at very low sequencing depths.

Finally, the lack of comprehensive data on cell composition as ground truth remains a recurrent limitation on the accurate resolution of TOO contributions from cfDNA data \cite{jackson2023deconvolution}. AE-based deconvolution has introduced substantial improvement in the field \cite{jackson2023deconvolution, wang2025cfdecon}, but performance at tumor fractions below 10\% remains insufficient for early-stage clinical use \cite{nguyen2024tissue}.
\subsubsection*{Lack of Standardization Across the cfDNA-Based MCED Pipeline: An Open Challenge}
Pre-analytical variables, including the blood draw-to-separation interval, freeze-thaw cycles, and collection tube type, introduce batch effects that are difficult to control across studies \cite{10.1093/clinchem/hvad194}. None of the reviewed methods addresses pre-analytical standardization, and without field-wide consensus on minimum input requirements and sample handling protocols, cross-study performance comparisons will remain confounded by uncontrolled upstream variation. Many existing methods, including, but not limited to, the reviewed methods, lack systematic demographic matching and adjustment \cite{bie2023multimodal, gao2022whole, annapragada2024genome, kim2024deep, shen2024development, zhang2025cfmethylpre}. Although certain characteristics of cfDNA (e.g., fragment size profiles) remain largely invariant between different demographics in healthy individuals \cite{aerden2026longitudinal}, variations in some variables (e.g., age and sex distribution) can still act as significant confounders in both methylation and fragmentomic analyses \cite{chen2025whole, 10.1093/clinchem/hvaf173, lee2026impact}. Even among healthy cohorts, research has demonstrated that these variables profoundly influence total cfDNA concentration and cell-type composition \cite{aerden2026longitudinal}. The best practice is to address any likely confounder that may inflate the results of a model. The consequence of not doing so is that the performance metrics reported are difficult to interpret as estimates of true clinical utility rather than cohort-specific artifacts. Differences in cohort composition, stage distribution, sequencing platform, and choice of performance metrics make it difficult to adequately aggregate evidence. For example, as there is no consensus on the definition of "early-stage" cancer and it can vary for each cancer type, the metrics should be reported for individual stages separately, optionally with an aggregated result across stages. In general, efforts must be made to establish minimal reporting standards for liquid biopsy studies (analogous to those adopted in genomics \cite{brazmaminimum}), as this is critical to accelerate the regulatory review and clinical deployment of cfDNA-based MCED assays.

\section{Conclusion}\label{conclusion}
The integration of fragmentomic and methylomic cfDNA signals with advanced computational modeling has reshaped the landscape of multicancer early detection, evolving from single marker detection to multimodal, interpretable, and cost-efficient frameworks capable of cancer detection even at low tumor fractions. Integrative frameworks demonstrate that methylation-fragmentation coupling can improve both detection sensitivity and TOO inference without necessarily increasing sequencing depth. Classical approaches continue to deliver robust, interpretable, and computationally efficient performance, and strong clinical translatability, although often at the cost of high sequencing depth, variable performance in CpG-sparse regions, or reliance on bisulfite treatment. Ensemble and hybrid deep learning frameworks currently offer the best overall balance of performance, feasibility at lower depths, and interpretability. That said, no single method represents a definitive approach, and future progress towards this goal will depend less on novel algorithms and architectures alone and more on the availability of large and demographically diverse cohorts, along with covariates--clinical and technical--being rigorously matched, outcomes stage-stratified, and external validation integrated in the test phase. In tandem, broader data availability is required to accelerate novelty in training models, in that, cfDNA datasets generated for clinical validation should, where ethically and legally permissible, be made available to the wider research community under FAIR (findable, accessible, interoperable, reusable) data principles, with infrastructures such as the EU's HealthData@EU under the European health data space (EHDS), which offers a model for federated, cross-border access to health data for secondary research use \cite{marcus2022european,hussein2024getting}. At the same time, given that genomic and methylation data could be re-identifiable, there is a serious concern under the general data protection regulation (GDPR). Addressing this demands a trusted research environment (TRE) that allows computation of data without compromising patient confidentiality.

Furthermore, adoption of a mandatory standardized framework for reporting performance metrics, preferably across stages, is an essential factor. Mandating these practices will facilitate reproducibility among methods and accelerate their regulatory and clinical translatability. Ultimately, cfDNA-based MCED assays that meet these standardized criteria form the most reliable foundation for large-scale real-world cancer screening programs.


\section{Conflicts of interest}
The authors declare that they have no competing interests.

\section{Funding}
Not applicable.

\section{Data availability}
Not applicable.

\section{CRediT authorship contribution statement }
\textcolor{black}{NS: Conceptualization, Formal analysis, Investigation, Writing – original draft, Writing – review and editing, Visualization, MW: Conceptualization, Supervision, Writing – review and editing, Resources, Funding acquisition, KR: Conceptualization, Supervision, Writing – review and editing, Resources, Funding acquisition.}

\section{Acknowledgments}
The authors thank anonymous reviewers for their valuable suggestions. 

\bibliographystyle{oup-plain}
\bibliography{bibliography}

\end{document}